\documentclass[10pt,twocolumn,letterpaper]{article}

\usepackage{cvpr}
\usepackage{times}
\usepackage{epsfig}
\usepackage{graphicx}
\usepackage{amsmath}
\usepackage{amssymb}
\usepackage{subfig}
\usepackage{eucal, url}
\usepackage{caption}
\cvprfinalcopy %

\begin{document}

\title{The Treasure beneath Convolutional Layers: \\
Cross-convolutional-layer Pooling for Image Classification\thanks{
This work is in part supported by ARC Grants FT120100969, and LP130100156.
Correspondence should be addressed to C. Shen (email: chhshen@gmail.com).
}
}

\author{
Lingqiao Liu$ ^1 $, Chunhua Shen$ ^{1,2} $, Anton van den Hengel$ ^{1,2} $\\
$ ^1 $ The University of Adelaide, Australia\\
$ ^2 $ Australian Centre for Robotic Vision
}

\maketitle
\begin{abstract}
A number of recent studies have shown that a Deep Convolutional Neural Network (DCNN) pretrained on a large dataset can be adopted as a universal image description which leads to astounding performance in many visual classification tasks. Most of these studies, if not all, adopt activations of the fully-connected layer of a DCNN as the image or region representation and it is believed that convolutional layer activations are less discriminative.

This paper, however, advocates that if used appropriately convolutional layer activations can be turned into a powerful image representation which enjoys many advantages over fully-connected layer activations. This is achieved by adopting a new technique proposed in this paper called cross-convolutional-layer pooling. More specifically, it extracts subarrays of feature maps of one convolutional layer as local features and pools the extracted features with the guidance of feature maps of the successive convolutional layer. Compared with exising methods that apply DCNNs in the local feature setting, the proposed method is significantly faster since it requires much fewer times of DCNN forward computation. Moreover, it avoids the domain mismatch issue which is usually encountered when applying fully connected layer activations to describe local regions.
By applying our method to four popular visual classification tasks, it is demonstrated that the proposed method can achieve comparable or in some cases significantly better performance than existing fully-connected layer based image representations while incurring much lower computational cost.

\end{abstract}

\tableofcontents

\clearpage
\section{Introduction}

Recently, Deep Convolutional Neural Networks (DCNNs) have attracted a lot of attention in visual recognition due to its good performance \cite{ImageNetDeepLearning}. It has been discovered that activations of a DCNN pretrained on a large dataset, such as ImageNet \cite{ImageNet}, can be employed as a universal image representation and applying this representation to many visual classification problems leads to astounding performance \cite{CNN_Baseline,ArxivNewBaseline}. This discovery quickly sparks a lot of interest and inspires a number of further extensions \cite{CNN_Regional,Our_NIPS}. A fundamental issue of this kind of methods is that how to generate image representation from a pretrained DCNN. Most of current solutions, if not all, take activations of the fully connected layer as the image representation. In contrast, activations of convolutional layers are rarely used and some studies \cite{VisualizeCNN,ExistingConvEx} have reported that directly using convolutional layer activations as image features produces inferior performance.

In this paper, however, we advocate that convolutional layer activations can be turned into a powerful image representation if they are used appropriately. We propose a new method called cross-convolutional layer pooling, or cross layer pooling in short, to fully leverage the discriminative information of convolutional layers.
The main contributions and also key differences to the previous attempt of using convolutional layer activations lay in two aspects: (1) we use convolutional layer activations in a `local feature' setting which extracts subarrays of convolutional layer activations as region descriptors. (2) we pool extracted local features with cross-layer information.

The first aspect is motivated by some recent works \cite{CNN_Regional,Our_NIPS} which have shown that DCNN activations are not translational invariant and it is beneficial to apply a DCNN to describe local regions and create the image representation by pooling multiple regional DCNN activations. Our method steps further to use subarrays of convolutional layer activations, that is, parts of CNN activations as regional descriptors. Compared with previous works \cite{CNN_Regional,Our_NIPS}, our method enjoys two major advantages: (1) instead of running DCNN forward computation multiple times, one for each local region, our method only needs to run a DCNN once (or very few times in our multiple-resolution scheme) for all local regions. This results in great computational cost saving. (2) existing methods \cite{CNN_Regional,Our_NIPS} essentially apply a network trained for representing an image to represent a local region. This causes significant domain mismatch between the input at the training stage and testing stages, which may degrade the discriminative power of DCNN activations. In contrast, our method avoids this issue since it still uses the whole image as the network input and only extracts parts of convolutional activations as regional descriptors.

The second aspect is motivated by the parts-based pooling methods \cite{zhangningpos,PANDA,ZhangNingECCV} which are commonly used in fine-grained image classification. This kind of methods create one pooling result for each detected part region and the final image representation is obtained by concatenating pooling results from multiple parts. We generalize this idea into the context of DCNN and avoid using any predefined parts annotation. More specifically, we deem the feature map of each filter in a convolutional layer as the detection response map of a part detector and apply the feature map to weight regional descriptors extracted from previous convolutional layer in the pooling process. The final image representation is obtained by concatenating pooling results from multiple channels with each channel corresponding to one feature map. Note that unlike existing regional-DCNN based methods \cite{CNN_Regional,Our_NIPS}, the proposed method does not need any additional dictionary learning and encoding steps at both training and testing stages. Besides the above two aspects, we also develop a simple scheme to extract local features at a finer resolution from convoluational layers and experiment with a coarse feature quantization scheme which significantly reduces the memory usage in storing image representations.

We conduct extensive experiments on four datasets covering four popular visual classification tasks, that is, scene classification, fine-grained object classification, generic object classification and attribute classification. Experimental results suggest that the proposed method can achieve comparable or in some cases significantly better performance than competitive methods while being considerably faster in creating image representations.

\begin{figure*}[ht!]
	\centering
    \includegraphics[height=60mm,width=100mm]{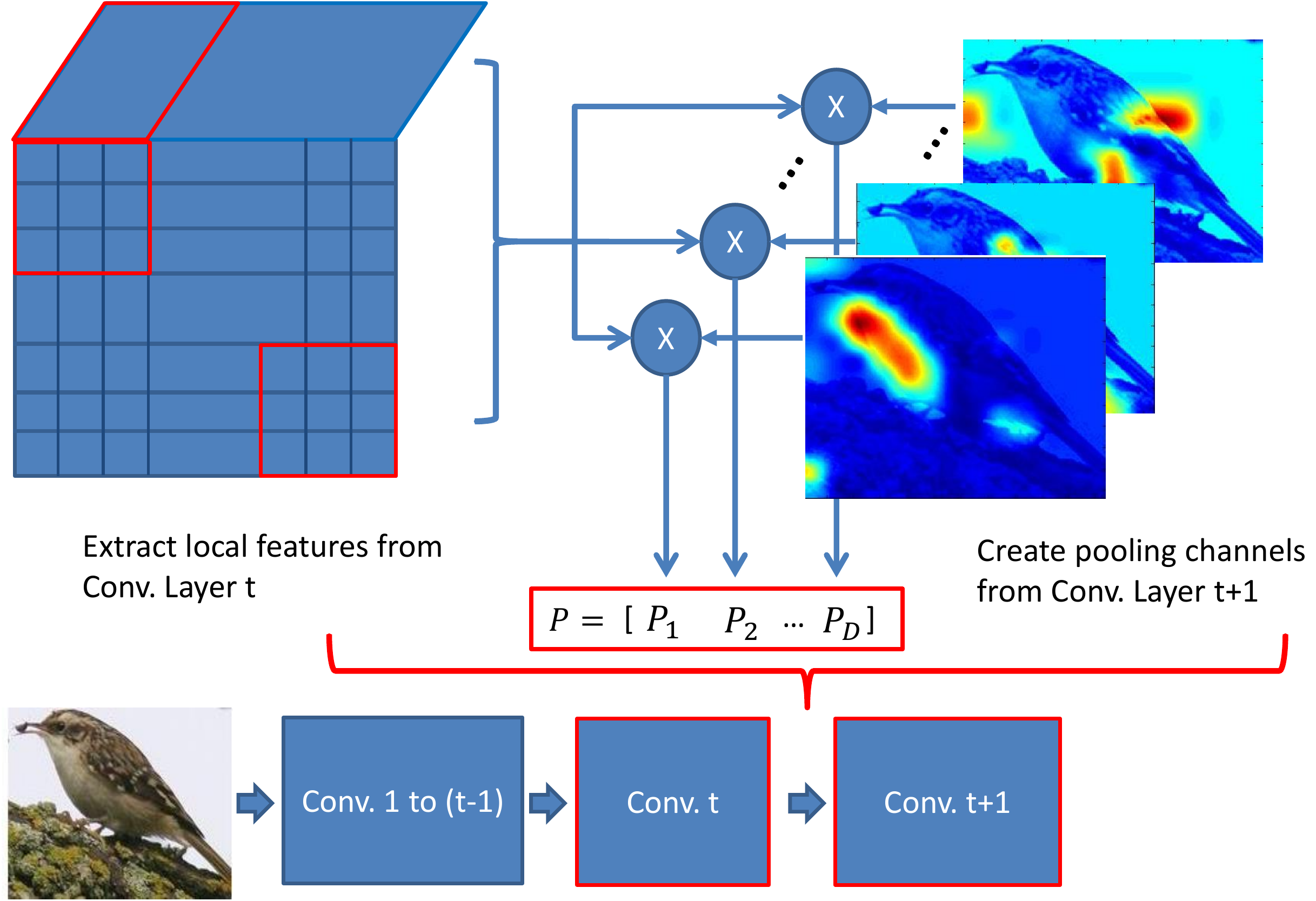}
    \caption{An overview of the proposed method.}
	\label{fig:overview}
\end{figure*}

\noindent \textbf{Preliminary:}
Our network structure and model parameters are identical to those in \cite{ImageNetDeepLearning}, that is, we have five convolutional layers and two fully connected layers. We use conv-1, conv-2, conv-3, conv-4, conv-5, fc-6, fc-7 to denote them respectively. At each convolutional layer, multiple filters are applied and it results in multiple feature maps, one for each filter. In this paper, we use the term `feature map' to indicate the convolutional results (after applying the ReLU) of one filter and the term `convolutional layer activations' to indicate feature maps of all filters in a convolutional layer.

\section{Current strategies for creating image representations from a pretrained DCNN}\label{sect:existing_ways}
In the literature, there are two major ways of using a pretrained DCNN to extract image representations: using a pretrained DCNN as global features and using a retrained DCNN as local features.

The first way takes the whole image as the input to a pretrained DCNN and extracts the fc-6/fc-7 activations as the image-level representation. To make the network better adapted to a given task, fine-tuning sometimes is applied. Also, to make this kind of methods more robust toward image transforms, averaging activations from several jittered versions of the original image, e.g. a slightly shifted version of the input image or a mirrored version of the input image, has been employed to obtain better classification performance \cite{ArxivNewBaseline}.

DCNNs can also be applied to extract local features. It has been suggested that DCNN activations are not invariant to a large amount of translation \cite{CNN_Regional} and the performance can be degraded if input images are not well aligned. To handle this, it is suggested to sample multiple regions from an input image and use one DCNN, called regional-DCNN in this scenario, to describe each region. The final image representation is aggregated from activations of those regional-DCNNs. Usually, another layer of unsupervised encoding is employed to create the image-level representation \cite{CNN_Regional, Our_NIPS}. Also, multiple-scale extraction strategy can be applied to further boost performance \cite{CNN_Regional}. It has been shown that for many visual tasks \cite{CNN_Regional,Our_NIPS} this kind of methods lead to better performance than directly extracting DCNN activations as global features.

One common factor of the above methods is that they all use fully-connnected layer activations as features. The convolutional layer activations, however, are not usually employed and some preliminary studies \cite{VisualizeCNN,ExistingConvEx} have suggested that the convolutional layer has weaker discriminative power than the fully-connected layer.

\section{Proposed method}

\begin{figure}
	\centering
    \includegraphics[height=35mm]{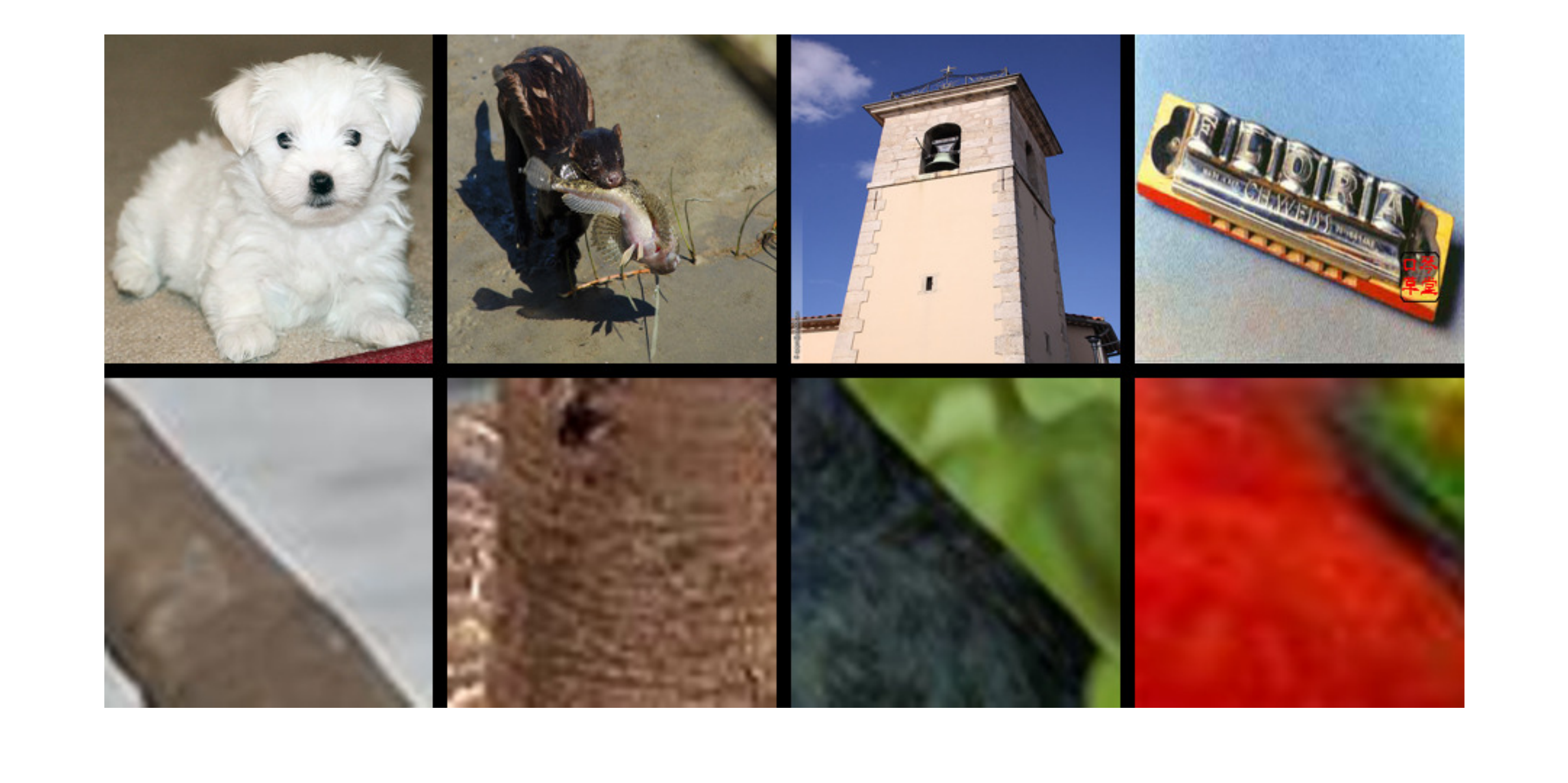}
    \caption{ This figure demonstrates the domain mismatch issue when applying the fully-connected layer activations as regional descriptors. Top row: input images that a DCNN `sees' at the training stage. Bottom row: input images that a DCNN `sees' at the test stage.}
	\label{fig:image_vs_region}
\end{figure}

\begin{figure}
	\centering
    \includegraphics[height=40mm]{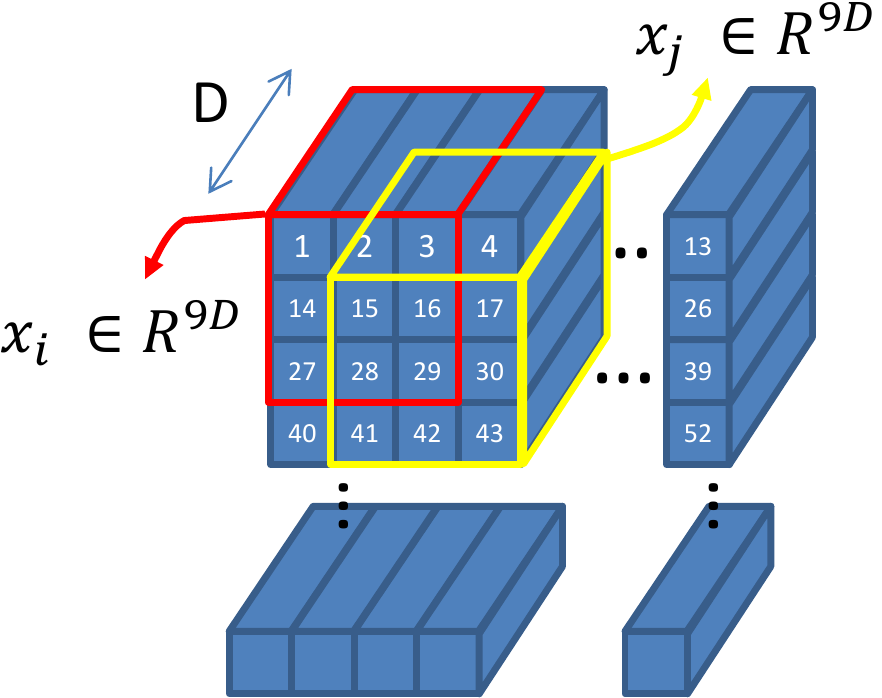}
    \caption{ The demonstration of extracting local features from a convolutional layer.}
	\label{fig:extract_conv_feature}
\end{figure}

\subsection{Convolutional layer vs. fully-connected layer}

One major difference between convolutional and fully-connected layer activations is that the former is embedded with rich spatial information while the latter does not. The convolutional layer activations can be formulated as a tensor of the size $H \times W \times D$, where $H,W$ denote the height and width of each feature map and $D$ denotes the number of feature maps. Essentially, the convolutional layer divides the input image into $H \times W$ regions and uses $D$-dimensional feature maps to describe the visual pattern within each region. Thus, convolutional layer activations can be viewed as a 2-D array of $D$-dimensional \textit{local features} with each one describing a local region. For the clarity of presentation, we call each of the $H \times W$ regions as a \textbf{spatial unit}, and the $D$-dimensional feature maps corresponding to a spatial unit as the \textbf{feature vector in a spatial unit}. The fully-connected layer takes the convolutional layer activations as the network input and transforms them into a feature vector representing the whole image. In this process, the spatial information is lost and the feature vector in a spatial unit cannot be explicitly recovered from activations of a fully-connected layer.

As mentioned in section \ref{sect:existing_ways}, DCNNs can also be applied to extract local features to handle the drawback of being translational variant. However, this scheme comes along with two side-effects in practice: (1) Its computational cost becomes higher than using DCNN activations as global image features since in this scheme one needs to run DCNN forward computing multiple times, one for each region. (2) Moreover, it makes the input of a DCNN become significant different from the input images that are used to train the network. This is because when applied as a regional feature, a DCNN is essentially used to describe local visual patterns which correspond to small parts of objects rather than whole images at the training stage. In Figure \ref{fig:image_vs_region}, we plot some training images from the ImageNet dataset and resized local regions. As can be seen, although they all have the same image size, their appearance and levels of details are quite different. Thus, blindly applying fully-connected layer activations as local features introduces significant domain mismatch which could potentially undermine the discriminative power of DCNN activations.

Our idea to handle aforementioned drawbacks is to extract multiple regional descriptors from \textit{a single DCNN applied to a whole image}. We realize this idea by leveraging the spatial information of convolutional layers. More specifically, in convolutional layers, we can easily locate a subset of activations which correspond to a local region. These subset of activations correspond to a set of subarrays of convolutional layer activations and we use them as local features. Figure \ref{fig:extract_conv_feature} demonstrates the extraction of such local features. For example, we can first extract $D$-dimensional feature vectors from regions $1,2,3,14,15,16,27,28,29$ and concatenate them into a $9\times D$-dimensional feature vector and then shift one unit along the horizontal direction to extract features from regions $2,3,4,15,16,17,28,29,30$. After scanning all the $13 \times 13$ feature maps we obtain 121 (omitting boundary spatial units) ($9\times D$)-dimensional local features.

It is clear that the proposed method enjoys the following merits: (1) the input of the DCNN is still a whole image rather than local regions. Thus the domain mismatch issue is avoided. (2) we only need to run DCNN forward calculation once and thus computational cost can be greatly saved. Note that in our method, we extract regional features from multiple spatial units and concatenate the feature vectors from them. This is different to previous work \cite{DNPRegionlets} (although their method is for a different application) which only treates the feature vector in one spatial unit as the local feature. We find that the use of feature vectors from multiple spatial units can significantly boost classification performance. This is because the feature vector from a single spatial unit may not be descriptive enough to characterize the visual pattern within a local region.

\subsection{Cross-convolutional-layer pooling}\label{sect:cl_pooling}

\begin{figure*}
  \centering
    \begin{tabular}{c}
            \subfloat{ \includegraphics[height=35mm]{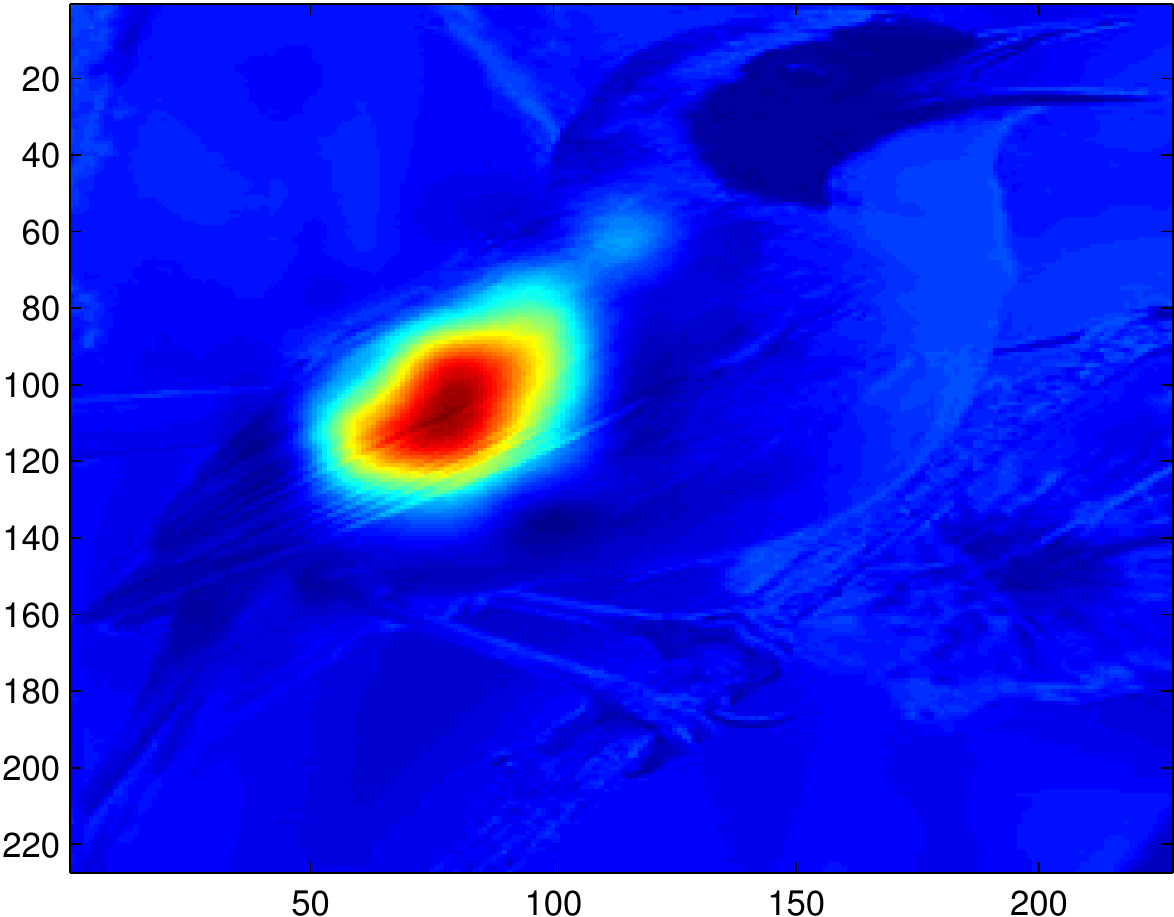}}
            \subfloat{ \includegraphics[height=35mm]{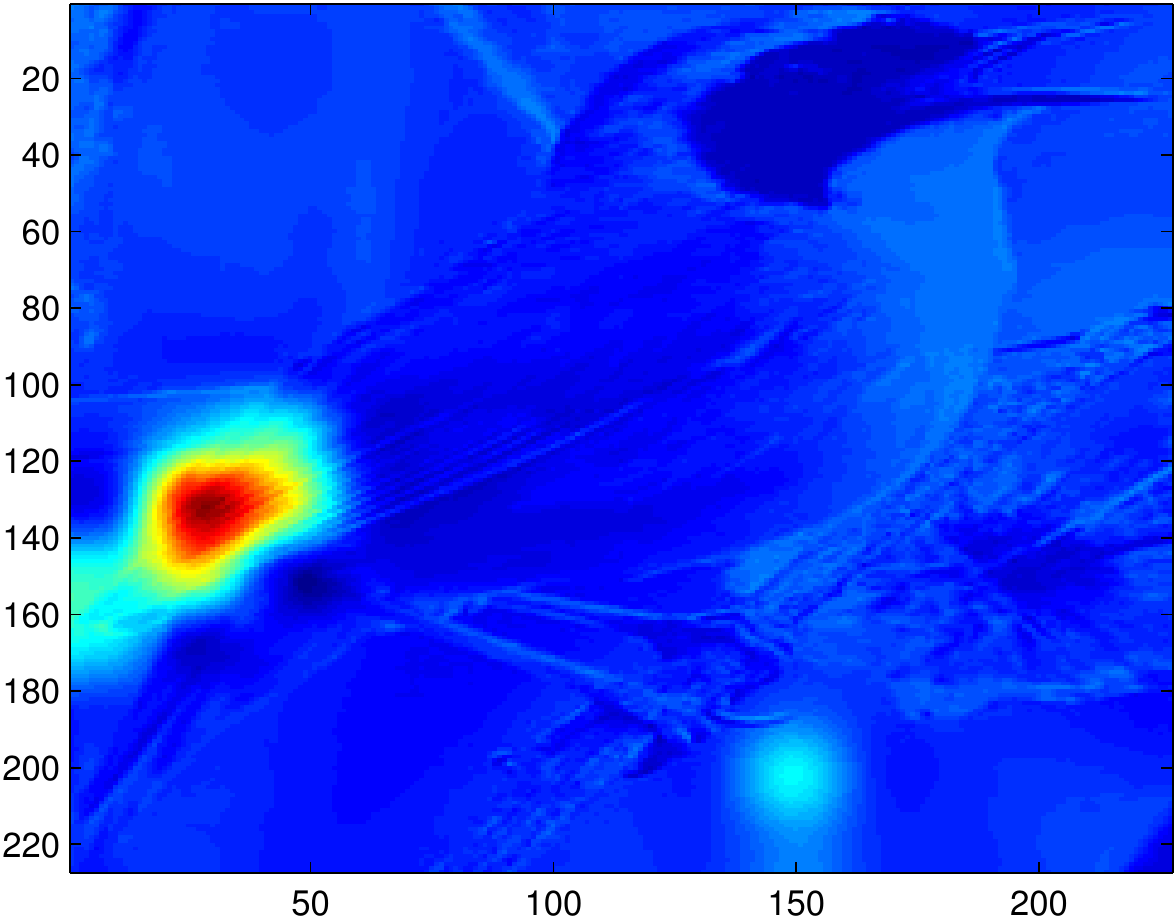}}
            \subfloat{ \includegraphics[height=35mm]{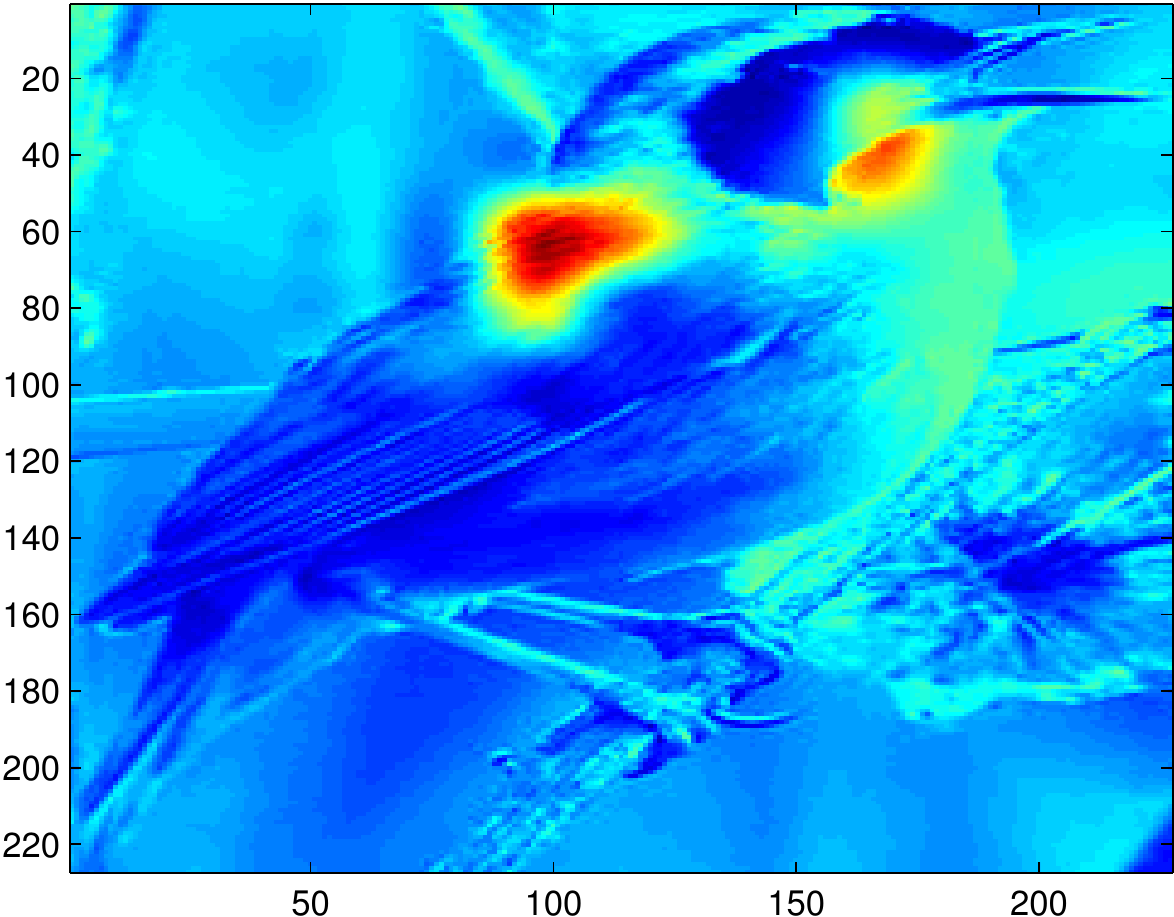}} \\
            \subfloat{ \includegraphics[height=35mm]{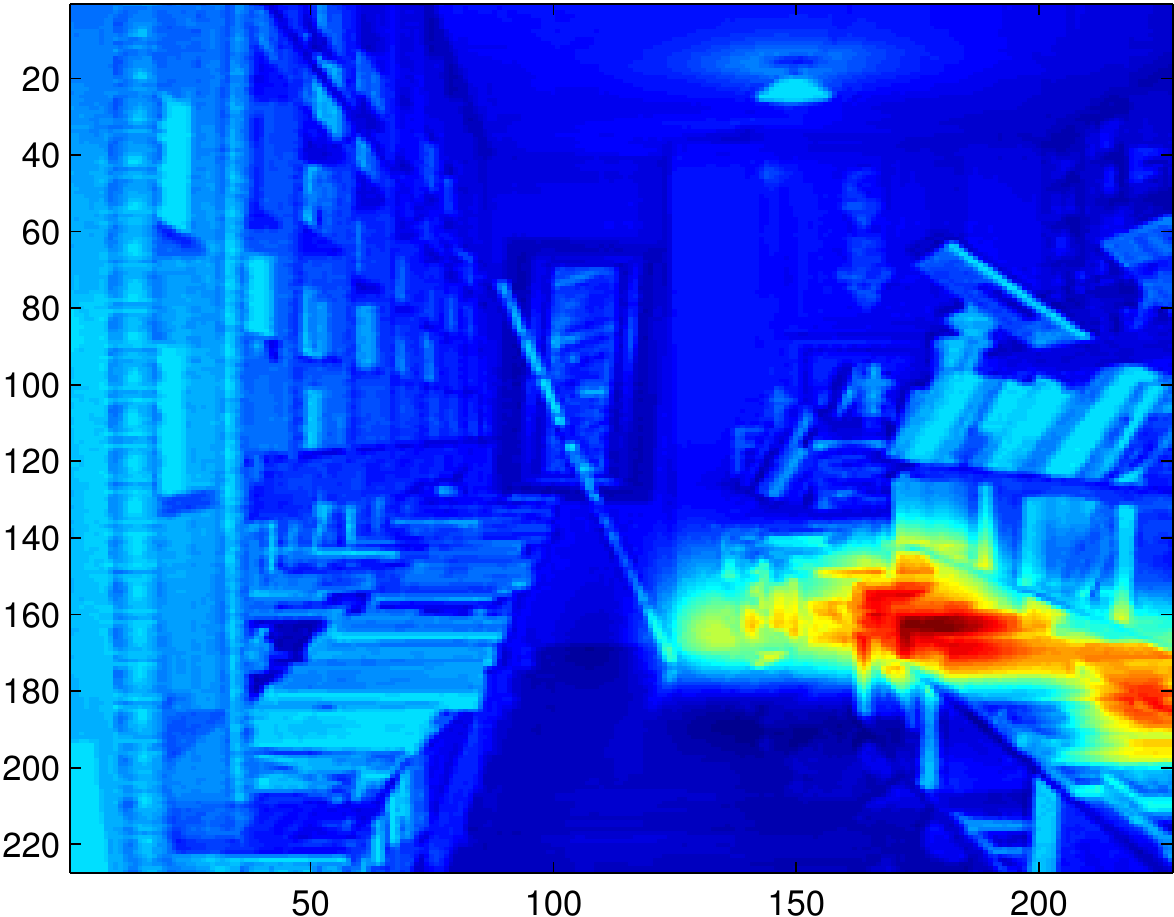}}
            \subfloat{ \includegraphics[height=35mm]{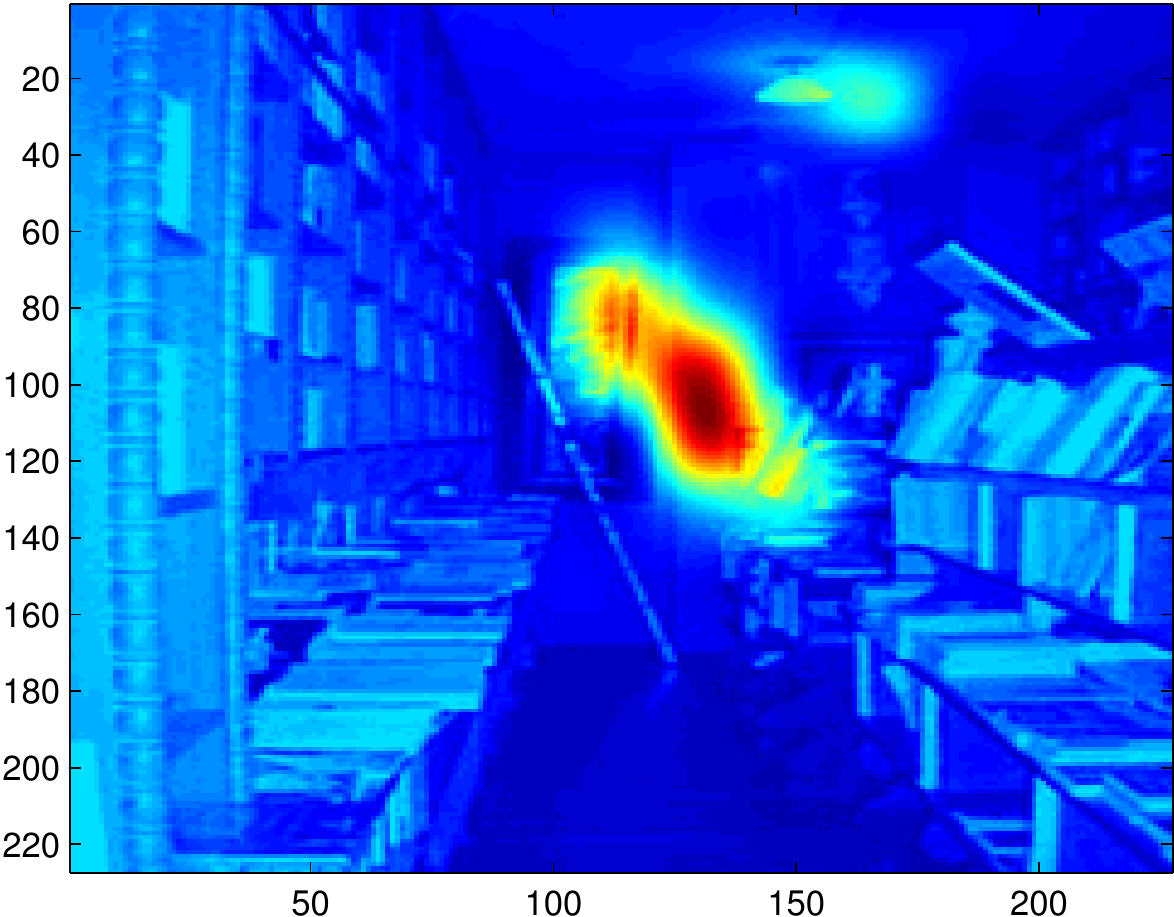}}
            \subfloat{ \includegraphics[height=35mm]{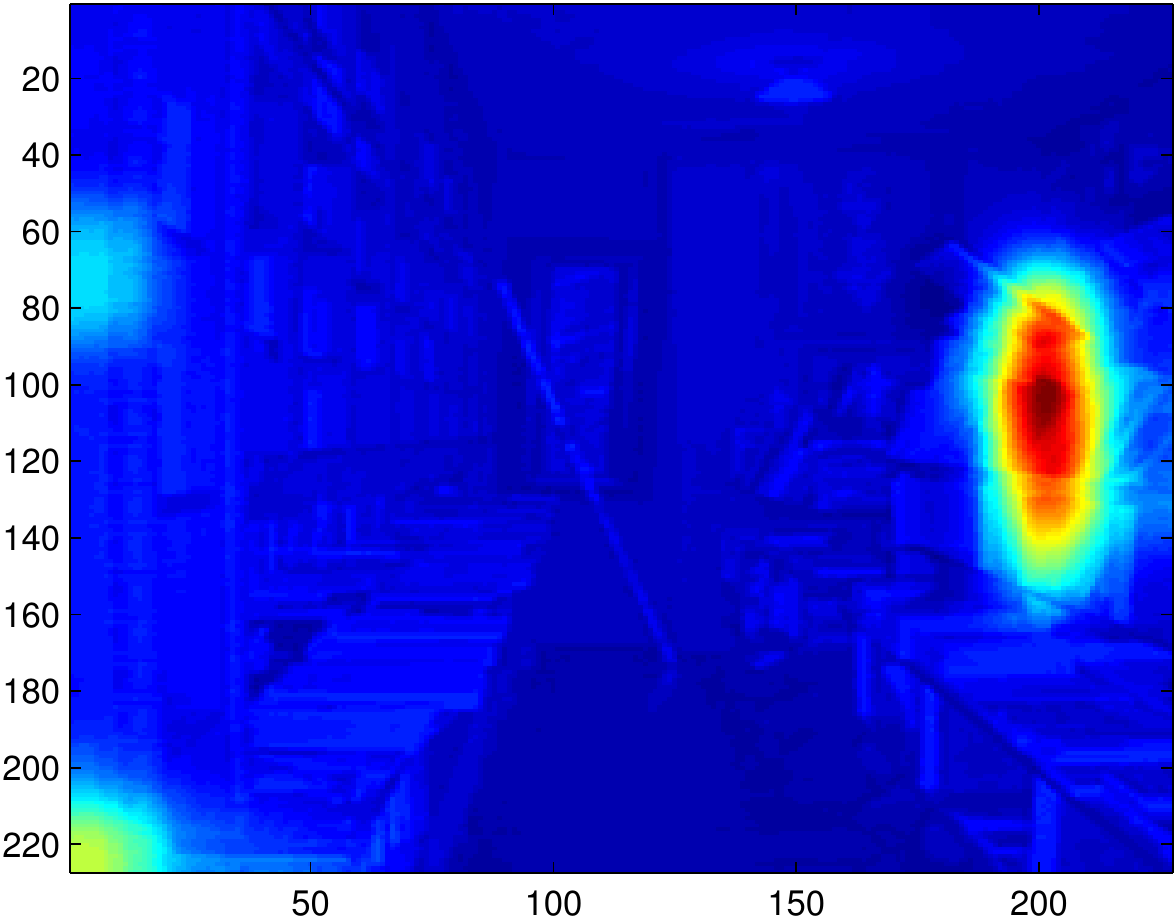}}
    \end{tabular}
    \caption{
          Visualization of some feature maps extracted from the 5th layer of a DCNN.
    }
    \label{fig:conv_visualize}
    \end{figure*}
After extracting local features from a convolutional layer, one can directly perform traditional max-pooling or sum-pooling to obtain the image-level representation. In this section, we propose an alternative pooling method which can significantly improve the classification performance. The proposed method is inspired by the parts based pooling strategy \cite{zhangningpos,PANDA,ZhangNingECCV} in fine-grained image classification. In this kind of methods, multiple regions-of-interest (ROI) are firstly detected and each of them corresponds to one human-specified object part, e.g. the tail of birds. Then local features falling into each ROI are pooled together to obtain a pooled feature vector. Given $D$ object parts, this strategy creates $D$ different pooled feature vectors and these vectors are concatenated together to form the final image representation. It has been shown that this simple strategy achieves significantly better performance than blindly pooling all local features together. Formally, the pooled feature from the $k$th ROI, denoted as $\mathbf{P}^{t}_k$, can be calculated by the following equation (let's consider sum-pooling in this case):
\begin{align}\label{Eq:part_pooling}
	\mathbf{P}^{t}_k = \sum_{i = 1} \mathbf{x}_i I_{i,k},
\end{align}
where $\mathbf{x}_i$ denotes the $i$th local feature and $I_{i,k}$ is a binary \textit{indicator map} indicating that if $\mathbf{x}_i$ falls into the $k$th ROI. We can also generalize $I_{i,k}$ to real value with its value indicating the `membership' of a local feature to a ROI. Essentially, each indicator map defines a pooling channel and the image representation is the concatenation of pooling results from multiple channels.

However, in a general image classification task, there is no human-specified parts annotation and even for many fine-grained image classification tasks, the annotation and detection of these parts are usually non-trivial. To handle this situation, in this paper we propose to use feature maps of the $(t+1)$th convolutional layer as $D_{t+1}$ indicator maps. By doing so, $D_{t+1}$ pooling channels are created for the local features extracted from the $t$th convolutional layer. We call this method cross-convolutional-layer pooling or cross-layer pooling in short. The use of feature maps as indicator maps is motivated by the observation that a feature map of a deep convolutional layer is usually sparse and indicates some semantically meaningful regions\footnote{Note that similar observation has also been made in \cite{VisualizeCNN}.}. This observation is illustrated in Figure \ref{fig:conv_visualize}. In Figure \ref{fig:conv_visualize}, we choose two images taken from two datasets, Birds-200 \cite{Birds200} and MIT-67 \cite{MIT67}. We randomly sample some feature maps from 256 feature maps in conv5 and overlay them to original images for better visualization. As can be seen from Figure \ref{fig:conv_visualize}, the activated regions of the sampled feature map (highlighted in warm color) are actually semantically meaningful. For example, the activated region in top-left corner of Figure \ref{fig:conv_visualize} corresponds to the wing-part of a bird.
Thus, the filter of a convolutional layer works as a part detector and its feature map serves a similar role as the part region indicator map. Certainly, compared with the parts detector learned from human specified part annotations, the filter of a convolutional layer is usually not directly task-relevant. However, the discriminative power of our image representation can be benefited from combining a much larger number of indicator maps, e.g. 256 as opposed to 20-30 (the number of parts usually defined by human), which is akin to applying bagging to boost the performance of multiple weak classifiers.

Formally, image representation extracted from cross-layer pooling can be expressed as follows:
\begin{align}
	& \mathbf{P}^{t} = [\mathbf{P}^{t}_1,\mathbf{P}^{t}_2,\cdots,\mathbf{P}^{t}_k,\cdots,\mathbf{P}^{t}_{D_{t+1}}] \nonumber \\
	& \mathrm{where,~~} \mathbf{P}^{t}_k = \sum_{i = 1}^{N_t} \mathbf{x}^{t}_i a^{t+1}_{i,k},
\end{align}
where $\mathbf{P}^{t}$ denotes the pooled feature for the $t$-th convolutional layer, which is calculated by concatenating the pooled feature of each pooling channel $\mathbf{P}^{t}_k, k = 1,\cdots,D_{t+1}$. $\mathbf{x}^{t}_i$ denotes the $i$th local feature in the $t$th convolutional layer. Note that feature maps of the $(t+1)$th convolutional layer is obtained by convolving feature maps of the $t$th convolutional layer with a $m\times n$-sized kernel. So if we extract local features $\mathbf{x}^{t}_i$ from each $m\times n$ spatial units in the $t$th convolutional layer then each $\mathbf{x}^{t}_i$ naturally corresponds to a spatial unit in the $(t+1)$th convolutional layer. Let's denote the feature vector in this spatial unit as $\mathbf{a}^{t+1}_{i} \in \mathbb{R}^{D_{t+1}}$ and the value at its $k$th dimension as $a^{t+1}_{i,k}$. Then we use $a^{t+1}_{i,k}$ to weight local feature $\mathbf{x}^{t}_i$ in the $k$th pooling channel.%

\noindent \textbf{Implementation Details:} In our implementation, we perform PCA on $\mathbf{x}^{t}_i$ to reduce the dimensionality of $\mathbf{P}^{t}$. Also, we apply power normalization to $\mathbf{P}^{t}$, that is, we use $\mathbf{\hat{P}}^{t} = \mathrm{sign}(\mathbf{P}^{t})\sqrt{|\mathbf{P}^{t}|}$ as the image representation to further improve performance. We also tried to directly use $\mathrm{sign}(\mathbf{P}^{t})$ as image representations, that is, we coarsely quantize $\mathbf{P}^{t}$ into $\{-1,1,0\}$ according to the feature sign of $\mathbf{P}^{t}$. To our surprise, this only introduces a slight performance drop. This observation allows us to simply use 2-bits to represent each feature dimension and this saves a lot of memory to store image representations. Please refer to section \ref{sect:coarse_quantization} for more detailed discussion.

\subsection{Creating finer resolutions of spatial units partitioning}\label{sect:multi-resolution}
\begin{figure*}
	\centering
    \includegraphics[height=60mm]{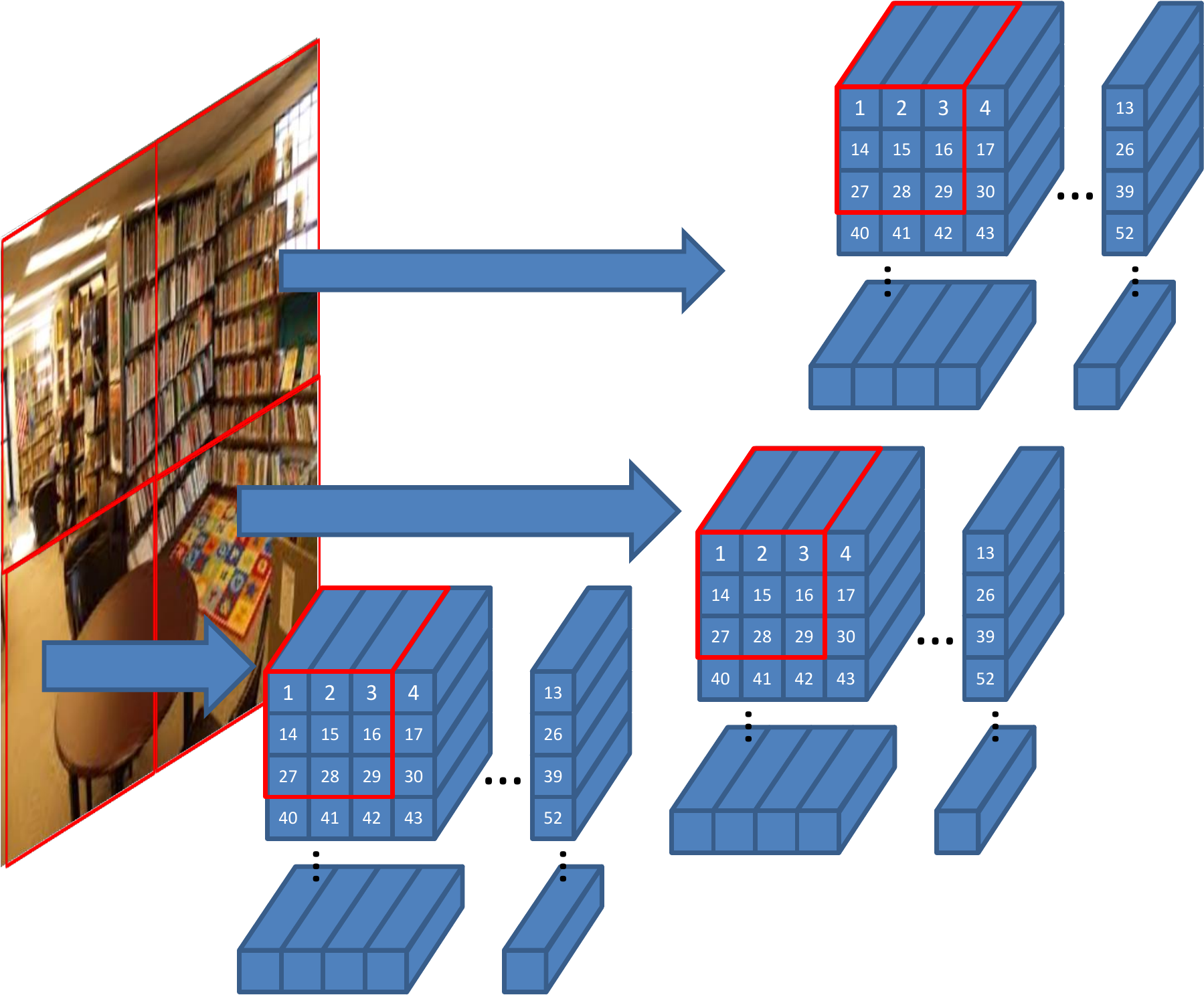}
    \caption{ Our scheme to create finer spatial unit partition.}
	\label{fig:finer_partition}
\end{figure*}

One drawback of the above method is that it only works in a single resolution. For example, if the 4-th and 5-th convolutional layer are used, the image can only be partitioned into $13\times 13$ spatial units. For some applications such as scene classification, the object of interest is usually small and it is favorable to use finer partitioning to capture finer object details. To achieve this goal, we proposed to divide the whole image into multiple non-overlap or overlapped blocks and apply a DCNN to each block. Then same as in section \ref{sect:cl_pooling}, local features are extracted from the convolutional layer of each DCNN. In total, it generates much more spatial units for the whole image, for example, if we partition the whole image into four quadrats we will have $26\times 26$ spatial units in total. This scheme is illustrated as in Figure \ref{fig:finer_partition}. Note that using this scheme we can easily create more spatial units by only introducing very few number of DCNN forward computation. In practice, we adopt a multi-resolution scheme which combines image representations extracted from multiple resolutions to achieve better classification performance.

\section{Experiments}
We evaluate the proposed method on four datasets: MIT indoor scene-67 (MIT-67 in short) \cite{MIT67}, Caltech-UCSD Birds-200-2011 \cite{Birds200} (Birds-200 in short), Pascal VOC 2007 \cite{pascal-voc-2007} (Pascal-07 in short) and H3D Human Attributes dataset \cite{HAT} (H3D in short). These four datasets cover several popular topics in image classification, that is, scene classification, fine-grained object classification, generic object classification and attribute classification. Previous studies \cite{CNN_Baseline,ArxivNewBaseline} have shown that using activations from the fc layer of a pretrained DCNN leads to surprisingly good performance in those datasets. Here in our experiments, we further compare different ways to extract features from a pretrained DCNN.
We organized our experiments into two parts, the first part compares the proposed method with other competitive methods and the second part examines the impact of various components in our method.

\subsection{Experimental protocol}

\begin{table}
        \caption{Comparison of results on MIT-67. The lower part of this table lists some results reported in the literature. The proposed methods are denote with *. For each method, the required number of CNN forward calculation is denoted as CNN $\times k$. }
		\centering
		\label{table:MIT67_Result}
    \scalebox{0.9}{

		\begin{tabular}{llll}
            \hline\noalign{\smallskip}
                Methods  &    Accuracy  & Comments \\
            \noalign{\smallskip}
            \hline
            \noalign{\smallskip}
			 CNN-Global				& 57.9\%          &   CNN$\times$1                \\
			 CNN-Jitter				& 61.1\%            &   CNN$\times$10                 \\
			 R-CNN SCFV	\cite{Our_NIPS} 			& 68.2\% 	        &   CNN$\times$100   \\
			 *CL-45         & 64.6\%           &   CNN$\times$1                 \\
			 *CL-45F        &  65.8\%          &   CNN$\times$4                  \\
			 *CL-45C       &  68.8\%          &   CNN$\times$5                  \\
			 *CL + CNN-Global   &  70.0\%          &   CNN$\times$6          \\
			 *CL + CNN-Jitter   &  \bf 71.5\%          &   CNN$\times$15             \\
            \noalign{\smallskip}
            \hline
            \noalign{\smallskip}
             Fine-tuning \cite{ArxivNewBaseline}                         & 66.0\%      & fine-tunning on MIT-67 \\
			 MOP-CNN    \cite{CNN_Regional}         		&  68.9\%	  & CNN$\times$53, three scales   \\
             VLAD level2  \cite{CNN_Regional}  			& 65.5\%          & CNN$\times$16, single scale  \\
			 CNN-SVM      \cite{CNN_Baseline}            & 58.4\%          & - \\
			 FV+DMS   \cite{Dis_Mode_Seeking}    	& 63.2\%		     & -  \\
			 DPM               \cite{DPM}      & 37.6\%	         & - \\
            \hline
      \end{tabular}
    }
    \end{table}

We compare the proposed method against three baselines, they are: (1) directly using fully-connected layer activations for the whole image (CNN-Global); (2) averaging fully-connected layer activations from several transformed versions of the input image. Following \cite{CNN_Baseline,ArxivNewBaseline}, we transform the input image by cropping its four corners and middle regions as well as by creating their mirrored versions; (3) the method in \cite{CNN_Baseline,ArxivNewBaseline} which extracts fully-connected layer CNN activations from multiple regions in an image and encodes them using sparse coding based Fisher vector encoding (RCNN-SCFV). Since RCNN-SCFV has demonstrated superior performance than the MOP method in \cite{CNN_Regional}, we do not include MOP in our comparison. To make fair comparison, we reimplement all three baseline methods. For the last method, we use the code provided by the author of \cite{Our_NIPS} to extract the regional CNN features and perform encoding.

For our method, we use the multi-resolution scheme suggested in section \ref{sect:multi-resolution}, that is, besides applying our method to the whole image, we also partition the image into $M \times N$ blocks and appy our method. The final image representation is the concatenation of image representations obtained from these two resolutions. For all datasets expect H3D, we set $M = N = 2$ and we set $ M = 2, N =1$ for H3D because most images in H3D have longer height than width.

In the first part of our experiments, we report the result obtained by using the 4th and 5th convolutional layer since using them achieves the best performance. We denote our methods as CL-45, CL-45F, CL-45C, corresponding to the settings of applying our method to the whole image, to multiple blocks for finer resolution and combining representations from two different resolutions respectively. We also conduct similar experiment on the 3-4th layer of a DCNN in the second part of experiments and denote them as CL-34, CL-34F and CL-34C.
To reduce the dimensionality of image representations, we perform PCA on local features extracted from convolutional layers and reduce their dimensionality to 500 before cross-layer pooling. In practice, we find that reducing to higher dimensionality only slightly increases the performance.
We use libsvm \cite{libsvm} as the SVM solver and use precomputed linear kernels as inputs. This is because the calculation of linear kernels/Gram matrices can be easily implemented in parallel. When feature dimensionality is high this part of computation actually occupies most of computational time. Thus it is appropriate to use parallel computing to accelerate this part.

\subsection{Performance evaluation}
\subsubsection{Classification results}
\noindent\textbf{Scene classification: MIT-67.}
MIT-67 is a commonly used benchmark for evaluating scene classification algorithms, it contains 6700 images with 67 indoor scene categories. Following the standard setting, we use 80 images in each category for training and 20 images for testing.  The results are shown in Table \ref{table:MIT67_Result}. It can be seen that all the variations of our method (methods with `*' mark in Table \ref{table:MIT67_Result}) outperforms the methods that use DCNN activations as global features (CNN-Global and CNN-Jitter). This clearly demonstrates the advantage of using DCNN activations as local features. We can also see that the performance obtained by combining CL-45 and CL-45F, denoted as CL-C, has already achieved the same performance as the regional-CNN based methods (R-CNN SCFV and MOP-CNN) while it requires much fewer times of CNN forward calculation. Moreover, combining with the global-CNN representation, our method can obtain further performance gain. By combining CL-C with CNN-Jitter, our method, denoted as CL+CNN-Global and CL+CNN-Jitter respectively, achieves impressive classification performance 71.5\%.

\noindent\textbf{Fine-grained image classification: Birds-200.}
Birds-200 is the most popular dataset in fine-grained image classification research. It contains 11788 images with 200 different bird species. This dataset provides ground-truth annotations of bounding boxes and parts of birds, e.g. the head and the tail, on both the training set and the test set. In this experiment, we just use the bounding box annotation. The results are shown in Table \ref{table:Birds_result}. As can be seen, the proposed method performs especially well on this dataset. Merely CL-45 has already achieved 72.4\% classification accuracy, wining 6\% improvement over the performance of R-CNN SCFV which as far as we know is the best performance obtained in the literature when no parts information is utilized. Combining with CL-45F, our performance can be improved to 73.5\%. It is quite close to the best performance obtained from the method that relies on strong parts annotation. Another interesting observation is that for this dataset, CL-45 significantly outperforms CL-45F, which is in contrary to the case in MIT-67. This suggests that the suitable resolution of spatial units may vary from dataset to dataset.

\begin{table*}
        \caption{Comparison of results on Birds-200. Note that the method with ``use parts'' mark requires parts annotations and detection while our methods do not employ these annotations so they are not directly comparable with us.}
		\centering
		\label{table:Birds_result}
		\begin{tabular}{llll}
            \hline\noalign{\smallskip}
                Methods  &    Accuracy & Remark \\
            \noalign{\smallskip}
            \hline
            \noalign{\smallskip}
			 CNN-Global				& 59.2\%          &   CNN$\times$1, no part.                \\
			 CNN-Jitter				& 60.5\%          &   CNN$\times$1, no part                 \\
			 R-CNN SCFV	\cite{Our_NIPS} 			& 66.4\% 	&   CNN$\times$100, no part   \\
			 *CL-45         & 72.4\%           &   CNN$\times$1, no part \\
			 *CL-45F        &  68.4\%          &   CNN$\times$4, no part            \\
			 *CL-45C       &  \bf 73.5\%          &   CNN$\times$5, no part \\
			 *CL + CNN-Global   &  72.4\%          &   CNN$\times$6, no part         \\
			 *CL + CNN-Jitter   &  73\%          &  CNN$\times$16, no part             \\

            \noalign{\smallskip}
            \hline
            \noalign{\smallskip}
                 GlobalCNN-FT \cite{ArxivNewBaseline}                  &  66.4 \% & no parts, fine tunning \\
            	 Parts-RCNN-FT    \cite{ZhangNingECCV}                  & 76.37 \% & use parts, fine tunning \\
            	 Parts-RCNN    \cite{ZhangNingECCV}                  & 68.7 \% & use parts, no fine tunning \\
			 CNNaug-SVM    \cite{CNN_Baseline}         	& 61.8\%	  &  CNN $\times$1 \\
             CNN-SVM  \cite{CNN_Baseline}  			    & 53.3\%     & CNN global \\
			 DPD+CNN \cite{Decaffe}   & 65.0\%          & use parts \\
			 DPD   \cite{Zhang_2013_ICCV}    	            & 51.0\%		     & -  \\
            \hline
      \end{tabular}
\end{table*}

\noindent\textbf{Object classification: Pascal-2007.}
Pascal VOC 2007 has 9963 images with 20 object categories. The task is to predict the presence of each object in each image. Note that most object categories in Pascal-2007 are also included in ImageNet. So ImageNet can be seen as a super-set of Pascal-2007. The results on this dataset are shown in Table \ref{table:Pascal_result}. From Table \ref{table:Pascal_result}, we can see that the best performance of our method (CL + CNN-Jitter) achieves comparable performance to the state-of-the-art. Also, by merely using the feature extracted from convolutional layer, our method CL-45C outperforms the CNN-Global and CNN-Jitter which use DCNNs to extract global image features. However, our CL-45C does not outperform R-CNN and our best performed method CL + CNN-Jitter does not achieve significant performance improvement as what it has achieved in MIT-67 and Birds-200. This is probably due to that the 1000 categories in ImageNet training set has already included 20 categories in Pascal-2007. Thus the fully-connected layer actually contains some classifier-level information and using fully-connected layer activations implicitly utilizes more training data from ImageNet.

\begin{table}
        \caption{Comparison of results on Pascal VOC 2007. }
		\centering
		\label{table:Pascal_result}
		\begin{tabular}{llll}
            \hline\noalign{\smallskip}
                Methods  &   mAP & Remark \\
            \noalign{\smallskip}
            \hline
            \noalign{\smallskip}
			 CNN-Global				& 71.7\%          &   CNN $\times$ 1                 \\
			 CNN-Jitter				& 75.0\%          &   CNN $\times$ 10                  \\
			 R-CNN SCFV	\cite{Our_NIPS} 			& 76.9\% 	&   CNN $\times$ 100   \\
			 *CL-45         & 72.6\%           &   CNN $\times$ 1                   \\
			 *CL-45F        &  71.3\%          &   CNN $\times$ 4                  \\
			 *CL-45C       &  75.0\%          &   CNN $\times$ 5                  \\
			 *CL + CNN-Global   &  76.5\%          &   CNN $\times$ 6           \\
			 *CL + CNN-Jitter   &  \bf 77.8\%          &   CNN $\times$ 15              \\
            \noalign{\smallskip}
            \hline
            \noalign{\smallskip}
			 CNNaug-SVM    \cite{CNN_Baseline}         	&  77.2\%	  & with augmented data  \\
             CNN-SVM      \cite{CNN_Baseline}  			& 73.9\%      & no augmented data  \\
			 NUS   \cite{NUS}    				& 70.5\%		     & -  \\
			 GHM   \cite{GHM}                   & 64.7\%	         & - \\
			 AGS  \cite{AGS}            		& 71.1\%           & - \\
            \hline
      \end{tabular}
\end{table}

\noindent\textbf{Attribute classification: H3D.}
In recent years, attributes of object, which are semantic or abstract qualities of object and can be shared by many categories, have gained increasing attention due to its potential use in zero/one-shot learning and image retrieval \cite{RelativeAttribute,RelativeAttributeSearch}. In this experiment, we evaluate the proposed method on the task of predicting attribute of human. We use H3D dataset \cite{HAT} which defines 9 attributes for a subset of `person' images from Pascal VOC 2007 and H3D. The results are shown in Table \ref{table:HAT_result}. Again, our method shows quite promising results. Merely using the information from convolutional layer, our approach has already achieved 77.3\% which outperforms R-CNN SCFV by 4\%. By combining with CNN-Jitter, our method becomes comparable to PANDA \cite{PANDA} which needs complicated poselet annotations and detections.

\begin{table}
        \caption{Comparison of results on Human attribute dataset. }
		\centering
		\label{table:HAT_result}
		\begin{tabular}{llll}
            \hline\noalign{\smallskip}
                Methods  &   mAP & Remark \\
            \noalign{\smallskip}
            \hline
            \noalign{\smallskip}
			 CNN-Global				& 74.1\%          &   CNN $\times$ 1                 \\
			 CNN-Jitter				& 74.6\%          &   CNN $\times$ 10                  \\
			 R-CNN SCFV	\cite{Our_NIPS} 			& 73.1\% 	&   CNN $\times$ 100   \\
			 *CL-45         & 75.3\%           &   CNN $\times$ 1                   \\
			 *CL-45F       &  70.7\%          &   CNN $\times$ 4                  \\
			 *CL-45C       &  77.3\%          &   CNN $\times$ 5                  \\
			 *CL + CNN-Global   &  78.1\%          &   CNN $\times$ 6           \\
			 *CL + CNN-Jitter   &  \bf 78.3\%          &   CNN $\times$ 15              \\
            \noalign{\smallskip}
            \hline
            \noalign{\smallskip}
             PANDA  \cite{PANDA}          &  78.9   & needs poselet annotation \\
             CNN-FT       \cite{ArxivNewBaseline}                   & 73.8 & CNN-Global, fine tunning \\
			 CNNaug-SVM    \cite{CNN_Baseline}         	&  73.0\%	  & with augmented data  \\
             CNN-SVM      \cite{CNN_Baseline}  			& 70.8\%      & no augmented data  \\
			 DPD   \cite{NUS}    				& 69.9\%		     & -  \\
            \hline
      \end{tabular}
\end{table}

\subsubsection{Computational cost}

\begin{table}
        \caption{Average time used for extracting an image representation for different methods. The time can be break down into two parts, time spend on extracting CNN features and time spend on performing pooling.}
		\centering
		\label{table:speed comparison}
		\begin{tabular}{lllll}
            \hline\noalign{\smallskip}
                Method  &    CNN Extraction & Pooling & Total \\
            \noalign{\smallskip}
            \hline
            		*CL-45        & 0.45s & 0.14s & 0.6s \\
   				*CL-45F		 & 1.3s  & 0.27s & 1.6s \\
   				*CL-45C       & 1.75s  & 0.41s & 2.2s \\
   				CNN-Global   & 0.4s  & 0s & 0.4s  \\
   				CNN-Jitter   & 1.8s  & 0s & 1.8s  \\
   				SCFV  \cite{Our_NIPS}       & 19s  & 0.3s & 19.3s \\
            \noalign{\smallskip}
            \hline
      \end{tabular}
\end{table}

It is clear that our method requires much less time in DCNN forward computation. To give an intuitive idea of the computational cost incurred by our method, we report the average time spend on extracting image representations of various methods in Table \ref{table:speed comparison}. As can bee seen, the computational cost of our method is comparable to that of
CNN-Global and CNN-Jitter. This is quite impressive given that our method achieves significantly better performance than these two methods. Compare with SCFV, the most competitive method to our approach in the sense of classification accuracy, we have obtained around 10 times speedup. Note that this speed evaluation is based our naive MATLAB implementation and our method can be further accelerated by using C++ or GPU implementation.

\subsection{Analysis of components of our method}
From the above experiments, the advantage of using the proposed method has been clearly demonstrated. In this section, we further examine the effect of various components in our method.

\subsubsection{Using different convolutional layers}
First, we are interested to examine the performance of using convolutional layers other than the 4th and 5th convolutional layers. We experiment with the 3th and 4th convolutional layers and report the resulting performance in Table \ref{table:3-4 Layer result}. From the result we can see that using 4-5th layers can achieve superior performance than 3-4th layers. This is not surprising since it has been observed that the deeper the convolutional layer is, the better discriminative power it has \cite{VisualizeCNN}.

\begin{table}
        \caption{Comparison of results obtained by using different pooling layers.}
		\centering
		\label{table:3-4 Layer result}
		\begin{tabular}{llllll}
            \hline\noalign{\smallskip}
                Method  &   MIT-67 & Birds200 & Pascal07 & H3D \\
            \noalign{\smallskip}
            \hline
            		CL-34        &   61.7\% & 64.6\%  & 66.3\% &  74.7\% \\
            		CL-34F       &   61.4\% & 61.4\%  & 64.9\% &  70.4\% \\
            		CL-34C     &  64.1\% & 66.8\% & 68.5\% & 75.9\% & \\
            		CL-45C       & \bf 68.8\%   & \bf 73.5\%  & \bf 75.0\% & \bf 77.3\%  \\
            \noalign{\smallskip}
            \hline
      \end{tabular}
\end{table}

\subsubsection{Comparison of different pooling schemes}
The cross-layer pooling is an essential component in our method. In this experiment, we compare it against other possible alternative pooling approaches, they are: directly performing sum-pooling (with square operation) and max-pooling, using spatial pyramid pooling as suggested in \cite{SPP_Conv}, applying the sparse coding based fisher vector encoding (SCFV) \cite{Our_NIPS} to encode extracted local features and perform pooling. To simplify the comparison, we only report results on the best performed single resolution setting for each dataset, that is, CL-45F for MIT-67 and CL-45 for the rest of three datasets. The results are shown in Table \ref{table:pooling_comparison}. As can be seen, the proposed cross-layer pooling significantly outperforms directly applying max-pooling or sum-pooling or even spatial-pyramid pooling. By applying another layer of encoding on local features before pooling, the classification accuracy can be greatly boosted. However, in most cases, its performance is still much inferior to the proposed method, as seen in cases of MIT-67, Pascal-07 and Birds-200. The only exception is the result on H3D, where SCFV performs slightly better than our method. However, it needs additional codebook learning and encoding computation while our method does not. Considering this computational benefit and superior performance in most cases, cross-layer pooling is clearly preferable than the other alternative methods.

\begin{table}
        \caption{Comparison of results obtained by using different pooling schemes}
		\centering
		\label{table:pooling_comparison}
		\begin{tabular}{llllll}
            \hline\noalign{\smallskip}
                Method  &   MIT-67 & Birds200 & Pascal07 & H3D \\
            \noalign{\smallskip}
            \hline
            		Direct Max            &  42.6\% & 52.7\% & 48.0\% & 61.1\% \\
            		Direct Sum-sqrt       &  48.4\% & 49.0\% & 51.3\% & 66.4\% \\
            		SPP  \cite{SPP_Conv}   &  56.3\% & 59.5\% & 67.3\% & 73.1\%  \\
            		SCFV \cite{Our_NIPS}     &  61.9\% & 64.7\% & 69.0\% & \bf 76.5\%  \\
            		CL-single & \bf 65.8\% & \bf 72.4\% & \bf 72.6\% & 75.3\% \\
            \noalign{\smallskip}
            \hline
      \end{tabular}
\end{table}

\subsubsection{Feature sign quantization}\label{sect:coarse_quantization}
Finally, we demonstrate the effect of applying a feature sign quantization to pooled feature. Feature sign quantization quantizes a feature to 1 if it is positive, -1 if it is negative and 0 if it equals to 0. In other words, we use 2 bits to represent each dimension of the pooled feature vector. This scheme greatly saves the memory usage. Similar to the above experiment setting, we only report the result on the best performed single resolution setting for each dataset. The results are shown in Table \ref{table:FS_Quantization}. Surprisingly, this coarse quantization scheme does not degrade the performance too much, in three datasets, MIT-67, Pascal-07 and H3D, it achieves almost same performance as the original feature.

\begin{table}
        \caption{Results obtained by using feature sign quantization.}
		\centering
		\label{table:FS_Quantization}
		\begin{tabular}{llll}
            \hline\noalign{\smallskip}
                Dataset  &   Feature sign quantization & Original \\
            \noalign{\smallskip}
            \hline
            		MIT-67            &  65.2\% & 65.8\% \\
            		Birds-200           & 71.1\%  & 72.4\% \\
            		Pascal07           &  71.2\% & 71.3\% \\
            		H3D    & 75.4\% & 75.3\% \\
            \noalign{\smallskip}
            \hline
      \end{tabular}
\end{table}

\section{Conclusion}

In this paper, we have proposed a new method called cross-convolutional layer pooling to create image representations from the convolutional activations of a pre-trained CNN. Through the extensive experiments, we have shown that this method enjoys good classification performance and low computational cost. Our discovery suggests that if used appropriately, convolutional layers of a pre-trained CNN contains very useful information and has many advantages over the scheme that using fully-connected layer activations as image representation.

\onecolumn
\bibliographystyle{ieee}
\bibliography{CSRef}

\end{document}